\documentclass[11pt,a4paper]{article}
\usepackage[hyperref]{acl2019}
\usepackage{times}
\usepackage{latexsym}
\usepackage{graphicx}
\usepackage{subcaption}
\usepackage{adjustbox}
\usepackage{url}

\aclfinalcopy % Uncomment this line for the final submission

\title{Enhancing PIO Element Detection in Medical Text Using Contextualized Embedding}

\author{Hichem Mezaoui \\
  IMRSV Data Labs \\ Ottawa, Canada \\
  \texttt{\small hichem@imrsv.ai} \\\And
  Aleksandr Gontcharov \\
  IMRSV Data Labs \\ Ottawa, Canada \\
  \texttt{\small aleksandr.gontcharov@imrsv.ai}\\\And
  Isuru Gunasekara \\
  IMRSV Data Labs \\ Ottawa, Canada \\
  \texttt{\small isuru@imrsv.ai}\\}

\date{}

\begin{document}
\maketitle
\begin{abstract}
 %Deep neural network based techniques are commonly used in Natural Language Processing (NLP) and have advanced the state of the art of text representation and information retrieval in general. Specifically, it enhanced the performance of several downstream NLP tasks through the use of contextualized embedding. Recently, such an embedding method, Bidirectional Encoder Representations from Transformers (BERT), further pushed the state of the art of several NLP tasks.
 
 In this paper, we investigate a new approach to Population, Intervention and Outcome (PIO) element detection, a common task in Evidence Based Medicine (EBM). 
 The purpose of this study is two-fold: to build a training dataset for PIO element detection with minimum redundancy and ambiguity and to investigate possible options in utilizing state of the art embedding methods for the task of PIO element detection. For the former purpose, we build a new and improved dataset by investigating the shortcomings of previously released datasets. For the latter purpose, we leverage the state of the art text embedding, Bidirectional   Encoder   Representations from Transformers (BERT), and build a multi-label classifier. We show that choosing a domain specific pre-trained embedding further optimizes the performance of the classifier.  Furthermore, we show that the model could be enhanced by using ensemble methods and boosting techniques provided that features are adequately chosen. 

\end{abstract}

\section{Introduction}

Evidence-based medicine (EBM) is of primary importance in the medical field. Its goal is to present statistical analyses of issues of clinical focus based on retrieving and  analyzing numerous papers in the medical literature \cite{haynes1997evidence}. The PubMed database is one of the most commonly used databases in EBM \cite{sackett1996evidence}.

Biomedical papers, describing randomized controlled trials in medical intervention, are published at a high rate every year. The volume of these publications makes it very challenging for physicians to find the best medical intervention for a given patient group and condition \cite{borah2017analysis}. Computational methods and natural language processing (NLP) could be adopted in order to expedite the process of biomedical evidence synthesis. Specifically, NLP tasks applied to well structured documents and queries can help physicians extract appropriate information to identify the best available evidence in the context of medical treatment.

Clinical questions are formed using the PIO framework, where clinical issues are broken down into four components: Population/Problem (P), Intervention (I), Comparator (C), and Outcome (O). We will refer to these categories as PIO elements, by using the common practice of merging the C and I categories. 
 In \cite{rathbone2017expediting} a literature screening performed in 10 systematic reviews was studied. It was found that using the PIO framework can significantly improve literature screening efficacy.  
Therefore, efficient extraction of PIO elements is a key feature of many EBM applications and could be thought of as a multi-label sentence classification problem.

Previous works on PIO element extraction focused on classical NLP methods, such as Naive Bayes (NB), Support Vector Machines (SVM) and Conditional Random Fields (CRF) \cite{chung2009sentence, boudin2010combining}. These models are shallow and limited in terms of modeling capacity. Furthermore, most of these classifiers are trained to extract PIO elements one by one which is sub-optimal since this approach does not allow the use of shared structure among the individual classifiers.

Deep neural network models have increased in popularity in the field of NLP. They have pushed the state of the art of text representation and information retrieval. More specifically, these techniques enhanced NLP algorithms through the use of contextualized text embeddings at word, sentence, and paragraph levels \cite{mikolov2013distributed,le2014distributed,peters2017semi,devlin2018bert,logeswaran2018efficient,radford2018improving}.

More recently, \newcite{jin2018pico} proposed a bidirectional long short term memory (LSTM) model to simultaneously extract PIO components from PubMed abstracts. To our knowledge, that study was the first in which a deep learning framework was used to extract PIO elements from PubMed abstracts.

In the present paper, we build a dataset of PIO elements by improving the methodology found in \cite{jin2018pico}. Furthermore,
we built a  multi-label PIO classifier, along with a boosting framework, based on the state of the art text embedding, BERT. This embedding model has been proven to offer a better contextualization compared to  a bidirectional LSTM model \cite{devlin2018bert}.  %Prior work has tackled the problem of biomedical evidence extraction  \cite{marshall2017automating,ferracane2016leveraging}. 

%\section{Experimental Setting}
\section{Datasets}

In this study, we introduce PICONET, a multi-label dataset consisting of sequences with labels  Population/Problem (P), Intervention (I),  and  Outcome  (O). This dataset was created by collecting structured abstracts from PubMed and carefully choosing abstract headings representative of the desired categories. The present approach is an improvement over a similar approach used in \cite{jin2018pico}.

Our aim was to perform automatic labeling while removing as much ambiguity as possible. We performed a search on April 11, 2019 on PubMed for 363,078 structured abstracts with the following filters: Article Types (Clinical Trial), Species (Humans), and Languages (English). Structured abstract sections from PubMed have labels such as introduction, goals, study design, findings, or discussion; however, the majority of these labels are not useful for P, I, and O extraction since most are general  (e.g. \textit{methods}) and do not isolate a specific P, I, O sequence. Therefore, in order to narrow down abstract sections that correspond to the P label, for example, we needed to find a subset of labels such as, but not limited to \textit{population}, \textit{patients}, and \textit{subjects}. We performed a lemmatization of the abstract section labels in order to cluster similar categories such as  \textit{subject} and \textit{subjects}. Using this approach, we carefully chose candidate labels for each P, I, and O, and manually looked at a small number of samples for each label to determine if text was representative.

Since our goal was to collect sequences that are uniquely representative of a description of Population, Intervention, and Outcome, we avoided a keyword-based approach such as in \cite{jin2018pico}. For example, using a keyword-based approach would yield a sequence labeled \textit{population and methods} with the label P, but such abstract sections were not purely about the population and contained information about the interventions and study design making them poor candidates for a P label. Thus, we were able to extract portions of abstracts pertaining to P, I, and O categories while minimizing ambiguity and redundancy. Moreover, in the dataset from \cite{jin2018pico}, a section labeled as P that contained more than one sentence would be split into multiple P sentences to be included in the dataset. We avoided this approach and kept the full abstract sections. The full abstracts were kept in conjunction with our belief that keeping the full section retains more feature-rich sequences for each sequence, and that individual sentences from long abstract sections can be poor candidates for the corresponding label.

For sections with labels such as \textit{population and intervention}, we created a mutli-label. We also included negative examples by taking sentences from sections with headings such as \textit{aim}. Furthermore, we cleaned the remaining data with various approaches including, but not limited to, language identification, removal of missing values, cleaning unicode characters, and filtering for sequences between 5 and 200 words, inclusive.
\begin{table}
\begin{center}
\begin{tabular}{ |p{3cm}|p{3cm}|p{3cm}|p{3cm}| } 
\hline
\textbf{Category} & \textbf{Sentences}  \\
\hline
 
I & 22818  \\ 
I O & 7  \\
I P & 337\\
O & 10994\\
P & 30106\\
P O & 13\\
NEGATIVE & 32053\\
\hline
\end{tabular}
\end{center}
\caption{Number of occurrences of each category P, I and O in abstracts.}
\end{table}
%\section{Classification Models}
%In the following sections we describe the models used to classify the elements of text and extract the PIO elements.

%\subsection{LSTM Model}
%In order to investigate the performance of a baseline model on the PICONET data, we fit an LSTM model.  The text was tokenized using the standard Keras tokenizer (Python library) with max vocabulary and fed through an embedding layer with dimension 100 and maximum sequence length of 200. The embedding layer was followed by an LSTM layer of dimension 32, and finally with a dense layer of dimension 3 and a sigmoid activation function (see Figure {\ref{fig:fig2}).

\section{BERT-Based Classification Model}
%\subsubsection{Background}
%In this work, we have built a classifier based on three embedding: SciBERT \cite{beltagy2019scibert}, BERT \cite{devlin2018bert} and BioBERT \cite{lee2019biobert}. \\
\subsection{Background}
BERT (Bidirectional Encoder Representations from Transformers) is a deep bidirectional attention text embedding model. The idea behind this model is to pre-train a bidirectional representation by jointly conditioning on both left and right contexts in all layers using  a transformer \cite{vaswani2017attention, devlin2018bert}. Like any other language model, BERT can be pre-trained on different contexts. A contextualized representation is generally optimized for downstream NLP tasks. 

Since its release, BERT has been pre-trained on a multitude of corpora. In the following, we describe different BERT embedding versions used for our classification problem.  The first version is based on the original BERT release \cite{devlin2018bert}. This model is pre-trained on the  BooksCorpus (800M words) \cite{zhu2015aligning} and English Wikipedia (2,500M words). For Wikipedia,  text passages were extracted while lists were ignored. The second version is BioBERT \cite{lee2019biobert}, which was trained on biomedical corpora: PubMed (4.5B words) and PMC (13.5B words). %The last version of embedding taken into account in this study is SciBERT \cite{beltagy2019scibert}. This model is pretrained on a random sample of 1.14M papers from Semantic Scholar \cite{ammar2018construction}. In the collected papers,$ 18 \%$ are from the computer science domain and $82 \%$ from broad biomedical domain. It is worth noting that the pre-training process was performed on the full texts and not only on abstracts. 

%To build the PIO classifier, we use three representation models:\\
%\textbf{BERT}:  This model pre-train deep bidirectional representations by jointly conditioning on both left and right context in all layers \cite{devlin2018bert};\\
%\textbf{SciBERT}: This model is based on BERT and trained on Biomedical and computer science corpora \cite{beltagy2019scibert};\\
%\textbf{BioBERT}: This model is also based on BERT and pretrained on biomedical corpora from PUBMED and PMC \cite{lee2019biobert}.
%\subsubsection{Classification Model}
\subsection{The Model}
The classification model is built on top of the BERT representation by adding a dense layer corresponding to the  multi-label classifier with three output neurons corresponding to PIO labels. In order to insure that independent probabilities are assigned to the labels,  as a loss function we have chosen the binary cross entropy with logits (BCEWithLogits) defined by
\begin{equation}
E=-\sum_{i=1}^{n} (t_i log(y_i)+(1-t_i)log(1-y_i));
\end{equation}
where \textbf{t} and \textbf{y} are the target and output vectors, respectively; \textbf{n} is the number of independent targets (n=3). The outputs are computed by applying the logistic function to the weighted sums of the last hidden layer activations, s,

\begin{equation}
y_i=\frac{1}{1+e^{-s_i}},
\end{equation}

\begin{equation}
s_i=\sum_{j=1}h_j w_{ji}.
\end{equation}

\begin{figure}[!ht]
\centering
\includegraphics[width=\linewidth]{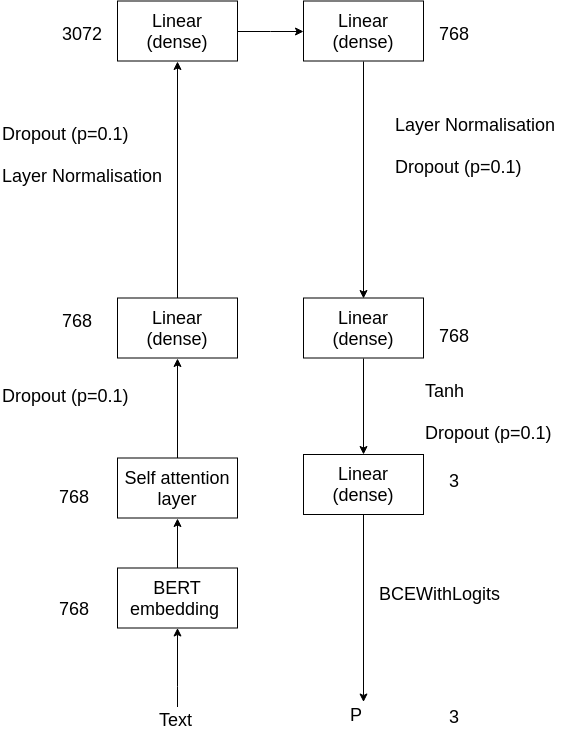}
\caption{Structure of the classifier.}
 
  \label{fig:fig2}
\end{figure}

 For the original BERT model, we have chosen the smallest uncased model, Bert-Base. The model has 12 attention layers and all texts are converted to lowercase by the tokenizer \cite{devlin2018bert}. The  architecture of the model is illustrated in Figure \ref{fig:fig2}.

Using this framework, we trained the model using the two pretrained embedding models described in the previous section. It is worth to mention that the embedding is contextualized during the training phase. For both models, the pretrained embedding layer is frozen during the first epoch (the embedding vectors are not updated). After the first epoch, the embedding layer is unfrozen and the vectors are fine-tuned for the classification task during training. The advantage of this approach is that few parameters need to be learned from scratch \cite{howard2018universal,radford2018improving,devlin2018bert}. 
 
\section{Results}
\subsection{Performance Comparison}
\begin{figure}[!ht]
  \centering
   \begin{subfigure}[b]{1.\linewidth}
    \includegraphics[width=\linewidth]{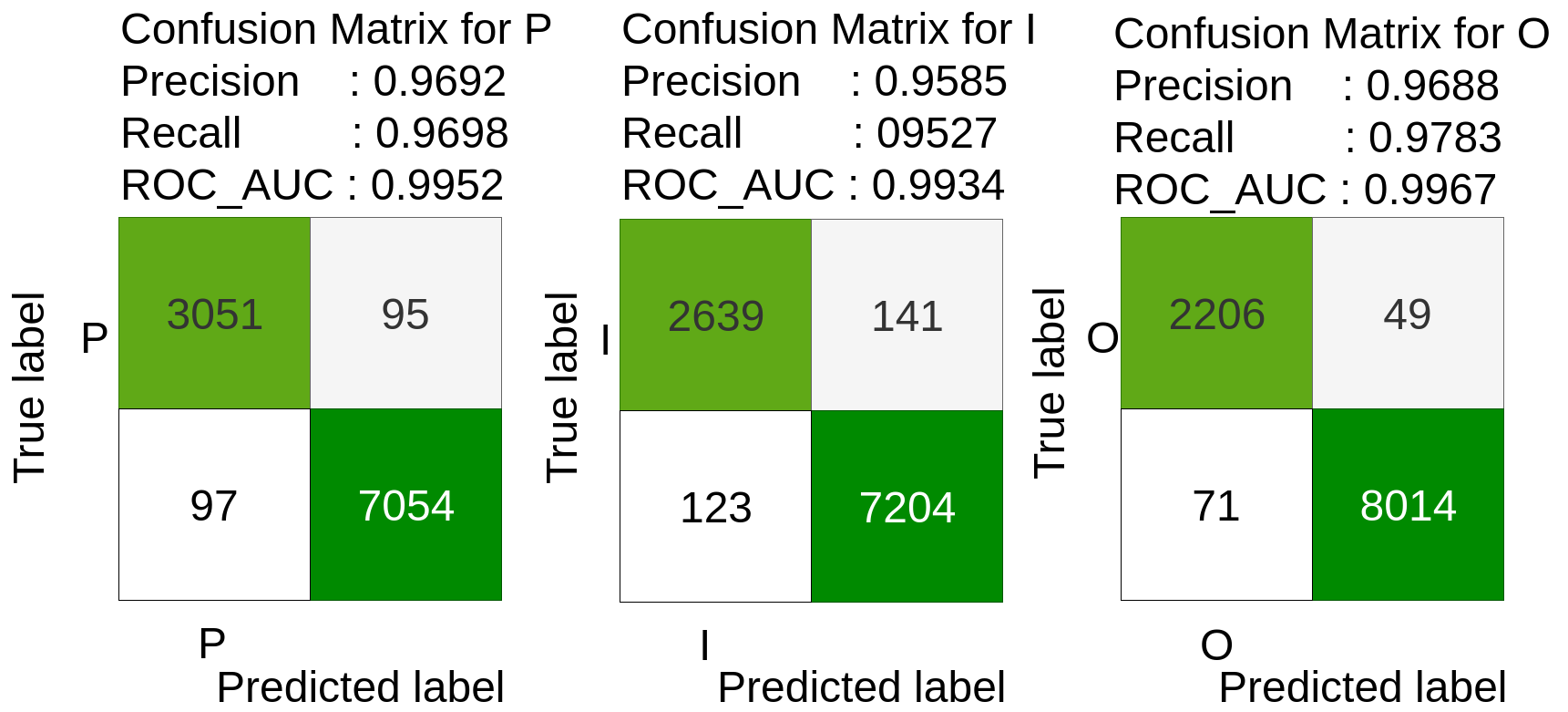}
    \caption{BERT (ROC\_AUC: 0.9951)}
  \end{subfigure}\vskip 1cm
  \begin{subfigure}[b]{1.\linewidth}
    \includegraphics[width=\linewidth]{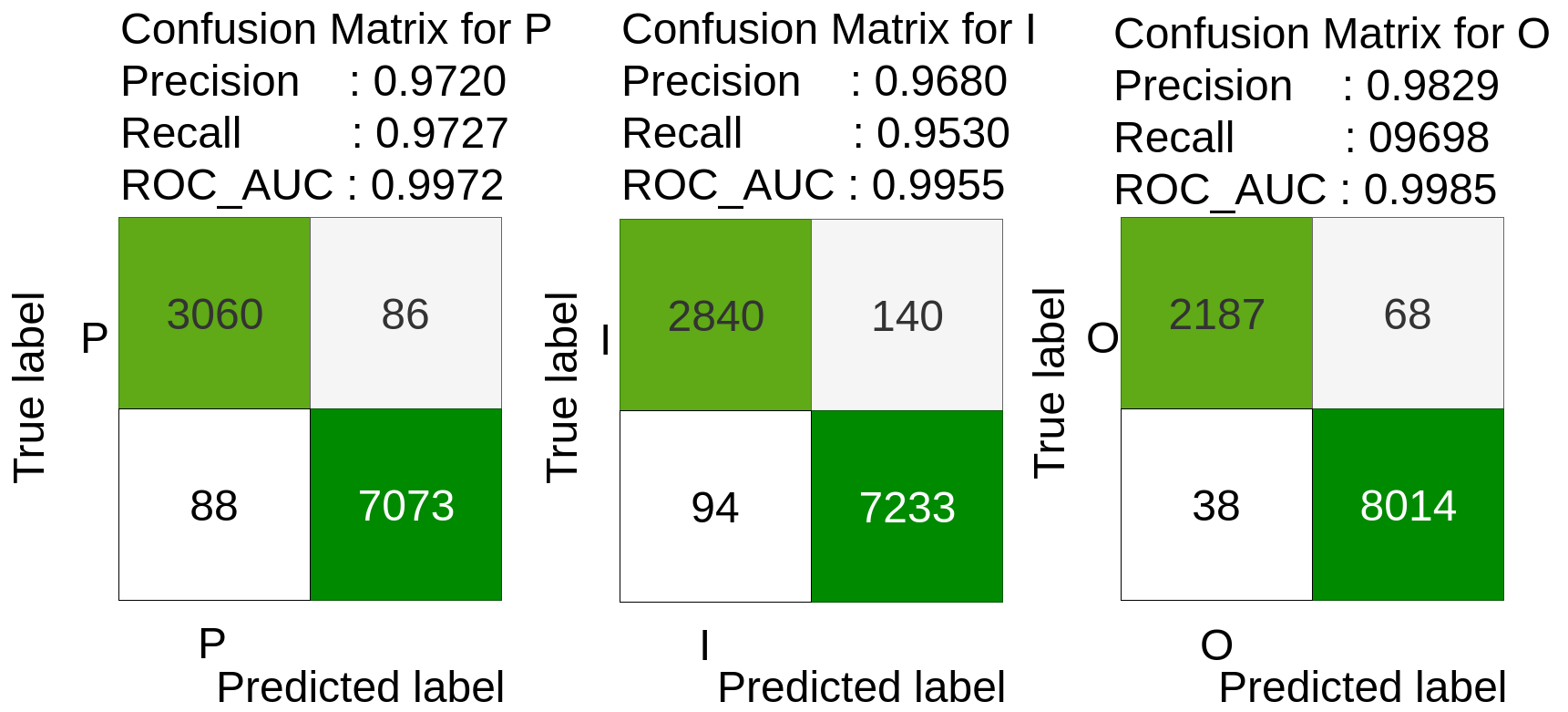}
    \caption{BioBERT (ROC\_AUC: 0.9971)}
  \end{subfigure}
  
 \caption{ROC\_AUC scores and confusion matrices.}
  \label{fig:fig1}
\end{figure}

In order to quantify the performance of the classification model, we computed the precision and recall scores. On average, it was found that the model leads to better results when  trained using the BioBERT embedding. In addition, the performance of the PIO classifier was measured by averaging the three Area Under Receiver Operating Characteristic Curve (ROC\_AUC) scores for P, I, and O. The ROC\_AUC score of 0.9951 was obtained by the model using the general BERT embedding. This score was improved to 0.9971 when using the BioBERT model pre-trained on medical context. The results are illustrated in Figure \ref{fig:fig1}.

\begin{figure}[!ht]
\centering
\includegraphics[width=\linewidth]{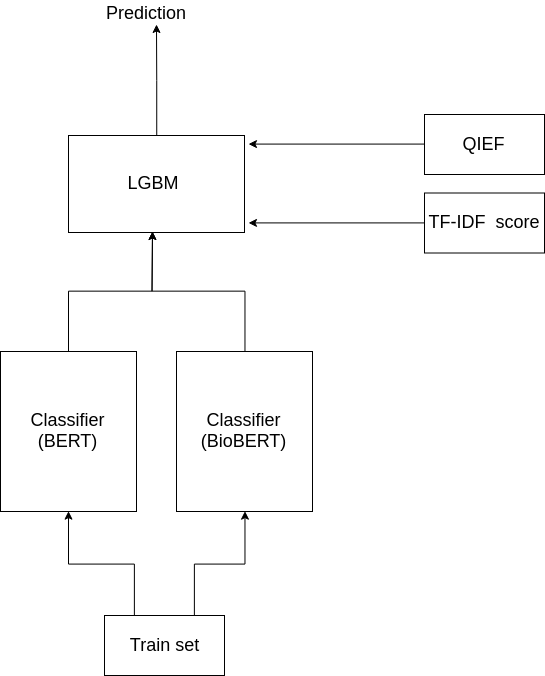}
\caption{An illustration of the LGBM framework: : combining the two base models and the TF-IDF and QIEF features.}
\end{figure}

\subsection{Model Boosting }

We further applied ensemble methods to enhance the model. This approach consists of combining predictions from base classifiers with features of the input data to increase the accuracy of the model \cite{merz1999using}. 

We investigate an important family of ensemble methods known as boosting, and more specifically a Light Gradient Boosting Machine (LGBM) algorithm, which consists of an implementation of fast gradient boosting on decision trees. In this study, we use a library implemented by Microsoft \cite{ke2017lightgbm}. In our model, we learn a linear combination of the prediction given by the base classifiers and the input text features to predict the labels. As features, we consider the average term frequency-inverse document frequency (TF-IDF) score for each instance and the  frequency of occurrence of quantitative information elements (QIEF) (e.g. percentage, population, dose of medicine). Finally, the output of the binary cross entropy with logits layer (predicted probabilities for the three classes) and the feature information are fed to the LGBM.

We train the base classifier using the original training dataset, using $60\%$ of the whole data as training dataset, and use a five-fold cross-validation framework to train the LGBM on the remaining $40\%$ of the data to avoid any information leakage. We train the LGBM on four folds and test on the excluded one and repeat the process for all five folds.

The results of the LGBM classifier for the different boosting frameworks and the scores from the base classifiers are illustrated in Table \ref{tab:boosting}. The highest average ROC\_AUC score of 0.9998 is obtained in the case of combining the two base learners along with the TF-IDF and QIEF features.% Whereas, the lowest ensemble AUC score of 0.9867, for the boosted models, is obtained using the LSTM model with the input text features.   \\
%\begin{figure}[h!]
%\centering
%\includegraphics[width=\linewidth]{bert_model.png}
%\caption{Structure of the neural network classifier (BERT)}
%\end{figure}

\begin{table}
\begin{center}
\scalebox{0.75}{
\begin{tabular}{ |c|c|c| }
\hline
\textbf{Model} & \textbf{ROC\_AUC} &\textbf{F1}  \\
\hline

BERT&0.9951&0.9666\\
 BioBERT &0.9971&0.9697\\
BERT + TF-IDF + QIEF & 0.9981&0.9784  \\ 
BioBERT + TF-IDF + QIEF & 0.9996 &0.9793 \\
BERT + BioBERT + TF-IDF + QIEF & 0.9998&0.9866\\

\hline
\end{tabular}}
\end{center}
\caption{Performance of the classifiers in terms of ROC\_AUC and F1 scores.}
\label{tab:boosting}
\end{table}

\section{Discussion and Conclusion}
In this paper, we presented an improved methodology to extract PIO  elements, with reduced ambiguity, from abstracts of medical papers. The proposed technique was used to build a dataset of PIO elements that we call PICONET. We further proposed a model of PIO elements classification using state of the art BERT embedding. It has been shown that using the contextualized BioBERT embedding improved the accuracy of the classifier. This result reinforces the idea of the importance of embedding contextualization in subsequent classification tasks in this specific context.

In order to enhance the accuracy of the model, we investigated an ensemble method based on the LGBM algorithm. We trained the LGBM model, with the above models as base learners, to optimize the classification by learning a linear combination of the predicted probabilities, for the three classes, with the TF-IDF  and QIEF scores. 
The results indicate that these text features were adequate for boosting the contextualized classification model. We compared the performance of the classifier when using the features with one of the base learners and the case where we combine the base learners along with the features. We obtained the best performance in the latter case.
%We obtained the highest score in terms of AUC when we combine the base learners. \\

The present work resulted in the creation of a PIO elements dataset, PICONET, and a classification tool. These constitute an important component of our system of automatic mining of medical abstracts. We intend to extend the dataset to full medical articles. The model will be modified to take into account the higher complexity of full text data and more efficient features for model boosting will be investigated.

%In order to enhance the accuracy of the model, we investigated an ensemble method based on the LGBM algorithm and shown that performance was optimized when the LGBM was trained on combining information about the predicted probabilities of the 3 classes from both models and TFIDF feature vectors in addition to the frequency of numerical entities in the text. Boosting the model and finding features of 

%\noindent \textbf{Preparing References:} \\
%Include your own bib file like this:
%\verb|\bibliographystyle{acl_natbib}|
%\verb|\bibliography{acl2019}| 

%where \verb|acl2019| corresponds to a acl2019.bib file.
\bibliography{acl2019}
\bibliographystyle{acl_natbib}

\appendix

\end{document}